\documentclass[12pt]{article}

\usepackage[T2A]{fontenc}
\usepackage[utf8]{inputenc}
 
\usepackage{hyphenat}
\usepackage{xcolor}
\usepackage{hyperref}
\definecolor{linkcolor}{HTML}{000000}
\definecolor{citecolor}{HTML}{0B6B1A}
\definecolor{urlcolor}{HTML}{0B2A5B}
\hypersetup{pdfstartview=FitH, citecolor=citecolor, linkcolor=linkcolor, urlcolor=urlcolor, colorlinks=true}

\usepackage[pdftex]{graphicx}
\graphicspath{{.\\}}
\DeclareGraphicsExtensions{.pdf,.png,.jpg}

\usepackage{authblk}

\usepackage{array}
\usepackage{appendix}

\begin{document}

\title{Correlation of the importances of neural network weights calculated by modern methods of overcoming catastrophic forgetting}
\author[ ]{Alexey Kutalev$^1$}
\affil[1]{EMLS Division, SberDevices, Sberbank PJSC, Moscow, Russia}
\affil[ ]{\it kutalev@gmail.com}

 
\date{\today}


\maketitle

\begin{abstract}

Following the invention in 2017 of the EWC method, several methods have been proposed to calculate the importance of neural network weights for use in the EWC method. Despite the significant difference in calculating the importance of weights, they all proved to be effective. Accordingly, a reasonable question arises as to how similar the importances of the weights calculated by different methods. To answer this question, we calculated layer-by-layer correlations of the importance of weights calculated by all those methods. As a result, it turned out that the importances of several of the methods correlated with each other quite strongly and we were able to present an explanation for such a correlation. At the same time, for other methods, the correlation can vary from strong on some layers of the network to negative on other layers. Which raises a reasonable question: why, despite the very different calculation methods, all those importances allow EWC method to overcome the catastrophic forgetting of neural networks perfectly?

\vspace{10pt} \textit{\textbf{Keywords:}
catastrophic forgetting, elastic weight consolidation, EWC, importance correlation, neural network, continual learning, machine learning}

\end{abstract}

\section{Introduction}

In 2017, a group of scientists from DeepMind introduced \cite{c5} the method of Elastic Weight Consolidation (EWC). The method is intended for use in continual learning process, that is, when a neural network is trained on several datasets consequently. After training on each dataset in the queue, the importance is calculated for each weight of the neural network produced by the training process. And, further, these importances are used in a special regularizing component of the loss function, which prevents the skills acquired on previous datasets from being forgotten in the learning process on the current and subsequent datasets. In the original work \cite{c5}, it is proposed to use the diagonal elements of the Fisher information matrix as the importance of the weight of the neural network:
\begin{equation}
\label{equ1}
F_i(\theta^*) = \left( \frac{\partial \log p(A | \theta^* )}{\partial \theta_i}\right)^2,
\end{equation}
where $\theta^*$ -- set of neural network weights where importances are calculated, $\theta_i$ -- $i$-th weight, $p(A | \theta^* )$ -- the probability to obtain the output values from dataset $A$ when the input values of $A$ are given to neural network.

Some time later, papers \cite{c7}, \cite{c8} and \cite{c10} were published, in which alternative methods for calculating the importance of weights were proposed, which were significantly different from (\ref{equ1}). But, despite the differences, the importances calculated by those methods made it possible to overcome catastrophic forgetting with a similar quality compared to the importances calculated using (\ref{equ1}). As shown in \cite{c8} and \cite{c11}, MAS method from \cite{c8} gives the most optimal values for the importances of weights to preserve previous skills when used instead of $F_i$.

Accordingly, the question arises: how importances calculated by different methods correlate with each other? For verification, we calculated the layer-by-layer correlation of the importances calculated for the same neural network by mentioned methods.

\section{Description of the correlation calculation methodology used}

The importances of the neural network weights in the papers \cite{c5} (denote FIS), \cite{c8} (denote MAS) and \cite{c10} (denote SIG) are calculated after the completion of training neural network on the dataset. The importances \cite{c7} (denote SI) are calculated directly during the training of the neural network on the dataset. Therefore, we used the following methodology. For the experiments, we used a neural network with three fully connected layers with the number of neurons 300, 150 and 10 in each layer and with the activation function $leakyReLU$ on the first two layers and $Softmax$ on the output layer. We trained neural network sequentially on ten datasets obtained from the MNIST dataset by ten random permutations on inputs (permuted MNIST \cite{c4}, \cite{c18}). Training on each dataset was performed for 6 epochs, and the importances were simultaneously calculated using the SI method. After the completion of training on each dataset, the importances were calculated and accumulated using the FIS, MAS and SIG methods. In the regularizing component of the loss function we used the importances by the MAS method to overcome catastrophic forgetting as providing the best quality. The optimal hyperparameter $\lambda$ of the EWC method was also used there, which was found by grid search for these network architecture and datasets.

Next, we calculated correlations between importances of different types after training on each dataset and for each layer of the neural network.  Then we built the visualization of surface of these correlations depending on layer and dataset in training queue.

\newpage

\section{Visualization of correlations}

\subsection{Correlation between MAS and FIS importances}

\begin{figure}[!ht]
\centering
\includegraphics[width=1\textwidth]{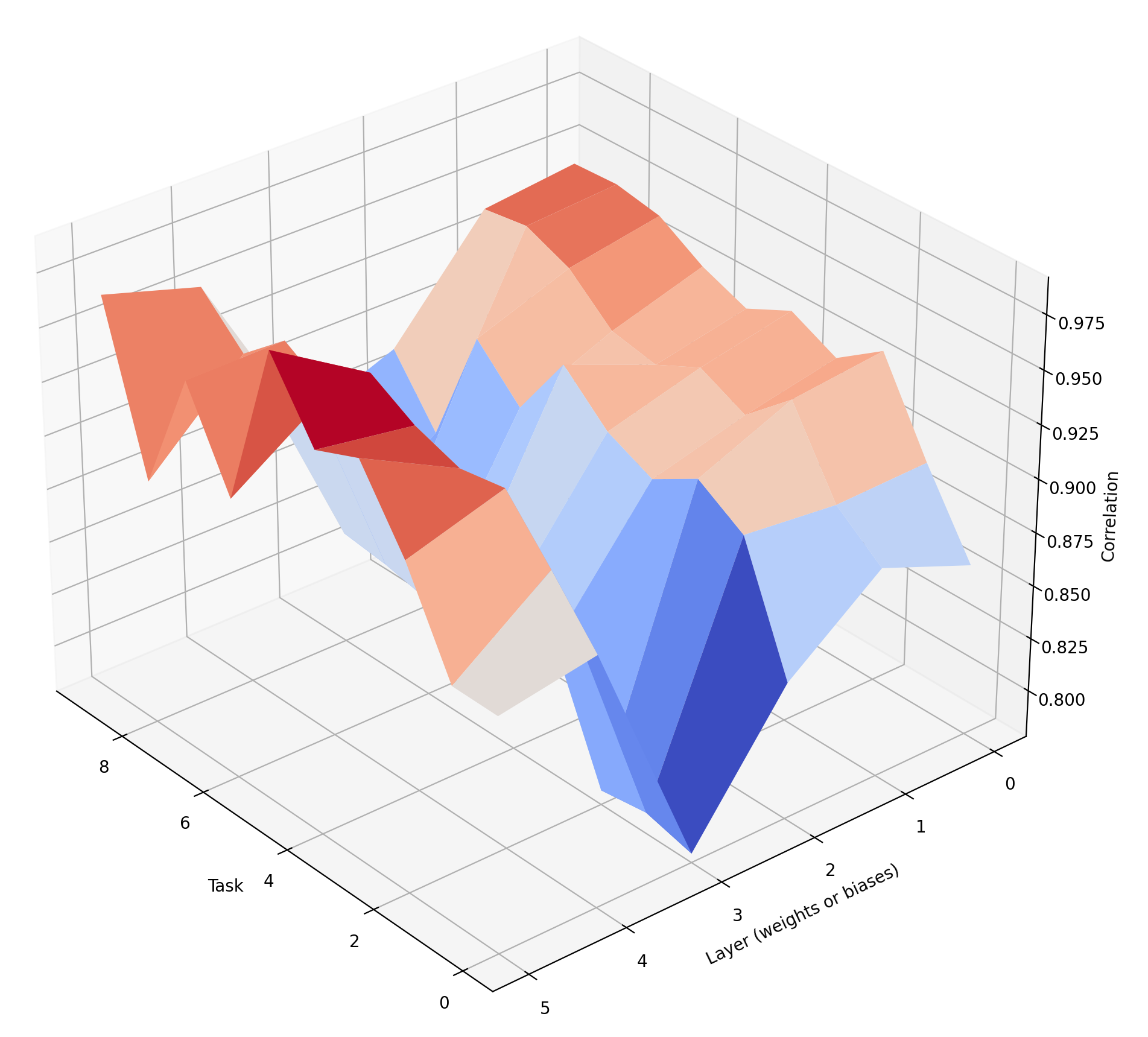}
\caption{\textit{Correlation between importances calculated by MAS and FIS methods.}}
\label{figure:1}
\end{figure}

Consider a graph in the form of a surface in Fig. \ref{figure:1}, reflecting the correlations between the importance of the weights calculated using the MAS and FIS methods. On one of the axes of the graph is a neural network layer, and on the other - the number of learned tasks (permuted MNIST datasets).

As we can see, correlation between MAS and FIS importances is very high -- more than 0.8. 

\begin{figure}[!ht]
\centering
\includegraphics[width=1\textwidth]{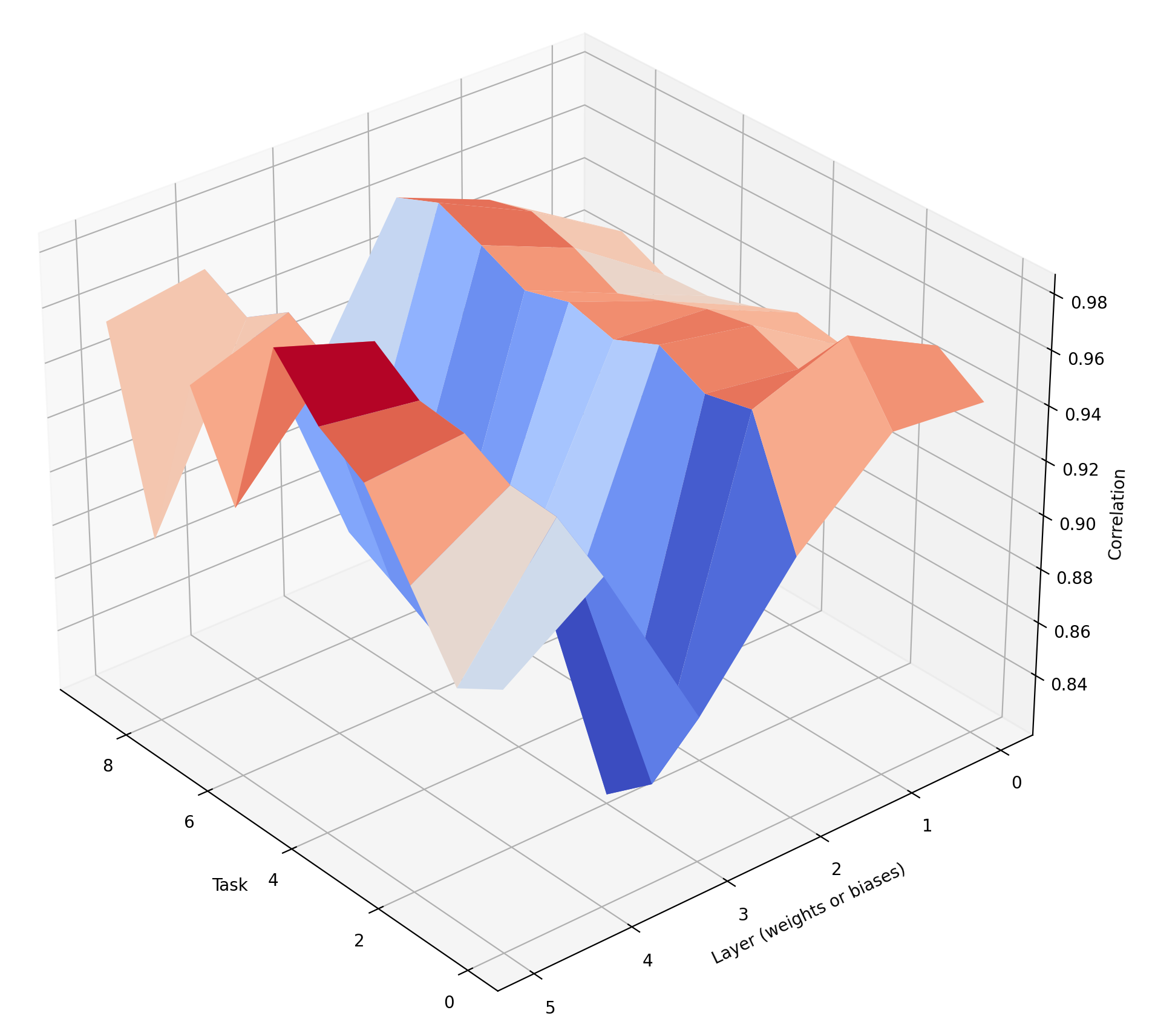}
\caption{\textit{Correlation between FIS and squared MAS importances.}}
\label{figure:7}
\end{figure}

Also we noted that correlation between FIS and squared MAS importances (see Fig. \ref{figure:7}) even more high than just FIS and MAS. We can propose the following explanation for this fact. The FIS importance can be approximated by the formula 

\begin{equation}
\label{equ2}
F_i(\theta^*) = \left( \frac{\partial \log p(A | \theta^* )}{\partial \theta_i}\right)^2 \approx \left( \frac{\partial \log \prod_k p(out_k | in_k ,  \theta^* ) }{\partial \theta_i} \right)^2 \approx
\end{equation}

$$
\approx \left( \sum_k \frac{1}{p_k(\theta^*)} \frac{ \partial p_k( \theta^* ) }{\partial \theta_i}\right)^2 \approx \left( \sum_k \frac{ \partial p_k( \theta^* ) }{\partial \theta_i}\right)^2,
$$
where $k$ runs over the samples of dataset, $p_k(\theta^*)$ is the labeled output of neural network with weights $\theta^*$ on $k$-th sample of dataset. Since the neural network has already trained on dataset then the outputs $p_k$ are very close to 1.

The MAS importance can be approximated by the formula 
\begin{equation}
\label{equ3}
\Omega_i(\theta^*) = \sum_k \frac{\partial \sum_j p_{kj}(\theta^* )^2}{\partial \theta_i} \approx 2 \sum_k \sum_j p_{kj} \frac{\partial  p_{kj}(\theta^* )}{\partial \theta_i} \approx  2 \sum_k \frac{\partial  p_{k}(\theta^* )}{\partial \theta_i}, 
\end{equation}
where $k$ runs over samples of dataset, $j$ runs over outputs of neural network, $p_{kj}$ is $j$-th output of neural network on $k$-sample of dataset. Since the neural network has already trained on dataset then the $p_{kj}$ is very close to 1 if $j$ is labeled output and close to 0 for other $j$.

The comparison between (\ref{equ2}) and (\ref{equ3}) may explain very high correlation between FIS and squared MAS importances. 

\newpage

\subsection{Correlation between MAS and SI importances}

\begin{figure}[!ht]
\centering
\includegraphics[width=1\textwidth]{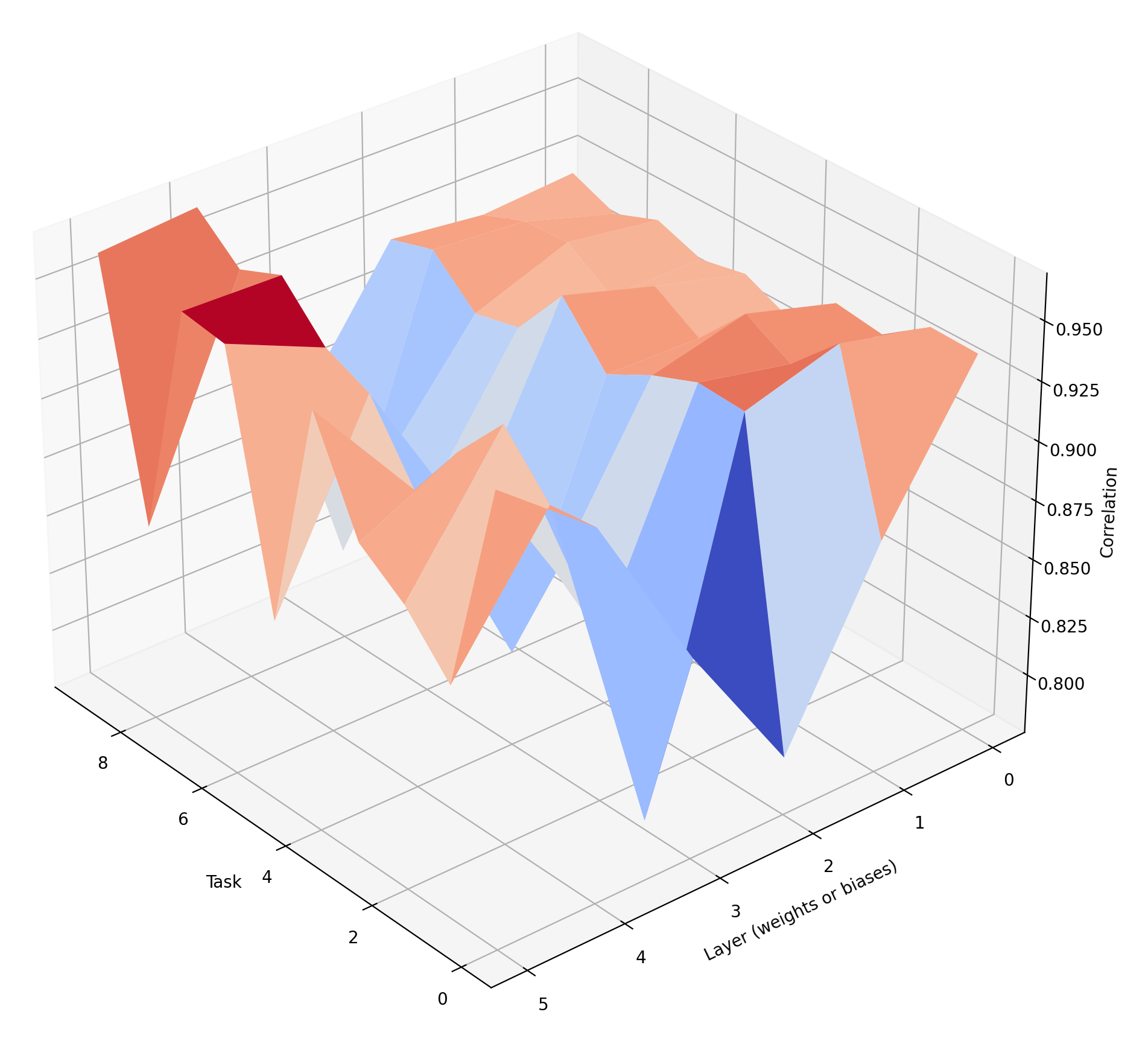}
\caption{\textit{Correlation between importances calculated by MAS and SI methods.}}
\label{figure:2}
\end{figure}

Now let's consider Fig. \ref{figure:2} -- graph of correlations between importances calculated by MAS and SI methods.

Correlation between MAS and FIS importances is also very high -- more than 0.8.

\subsection{Correlation between MAS and SIG importances}

\begin{figure}[!ht]
\centering
\includegraphics[width=1\textwidth]{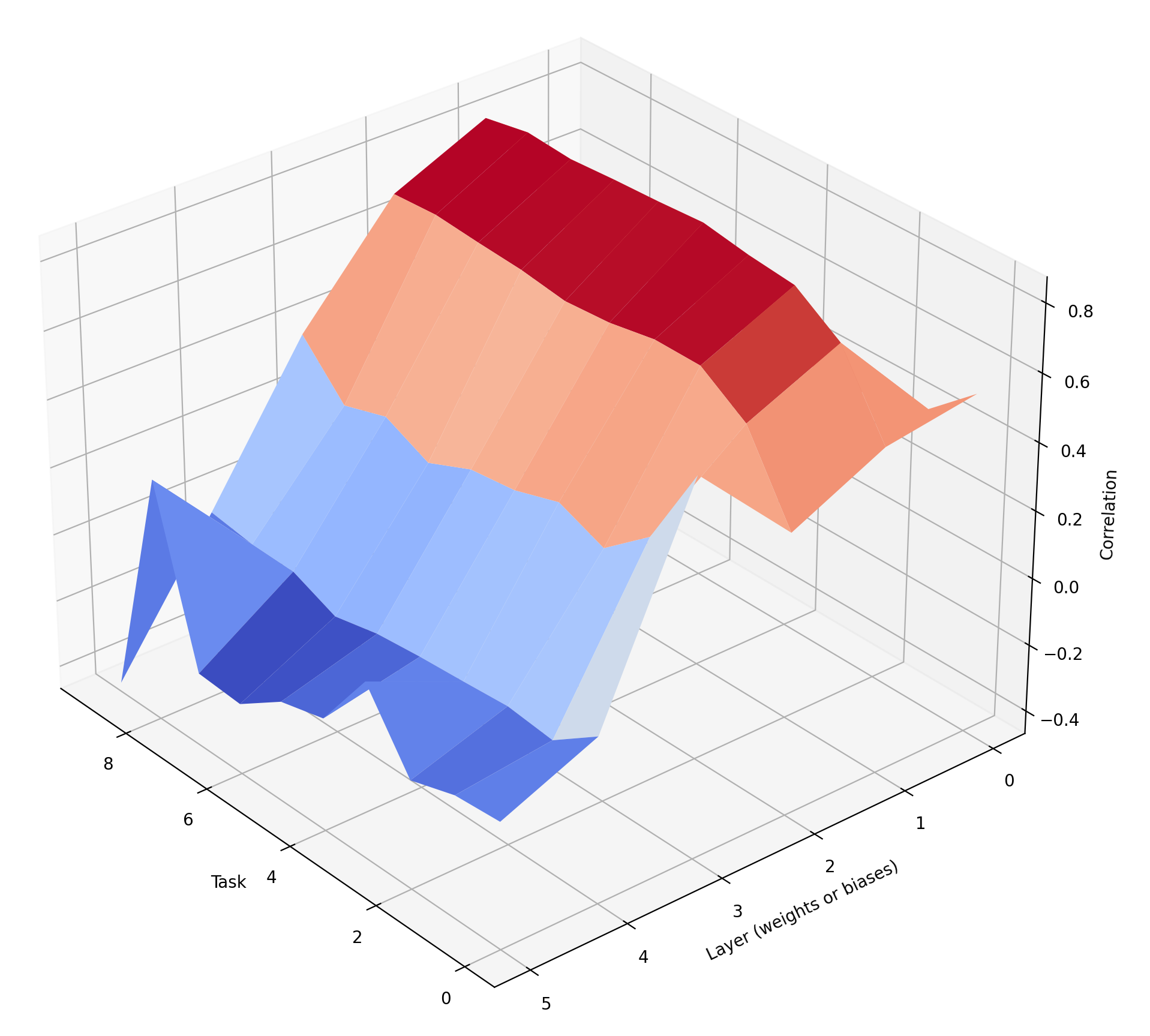}
\caption{\textit{Correlation between importances calculated by MAS and SIG methods.}}
\label{figure:3}
\end{figure}

Consider Fig. \ref{figure:3} -- graph of correlations between importances calculated by MAS and SIG methods.

Correlation between MAS and SIG importances is very different  -- from high (about 0.6) on input layers of neural network, to negative on output layer.

\subsection{Correlation between FIS and SI importances}

\begin{figure}[h!]
\centering
\includegraphics[width=1\textwidth]{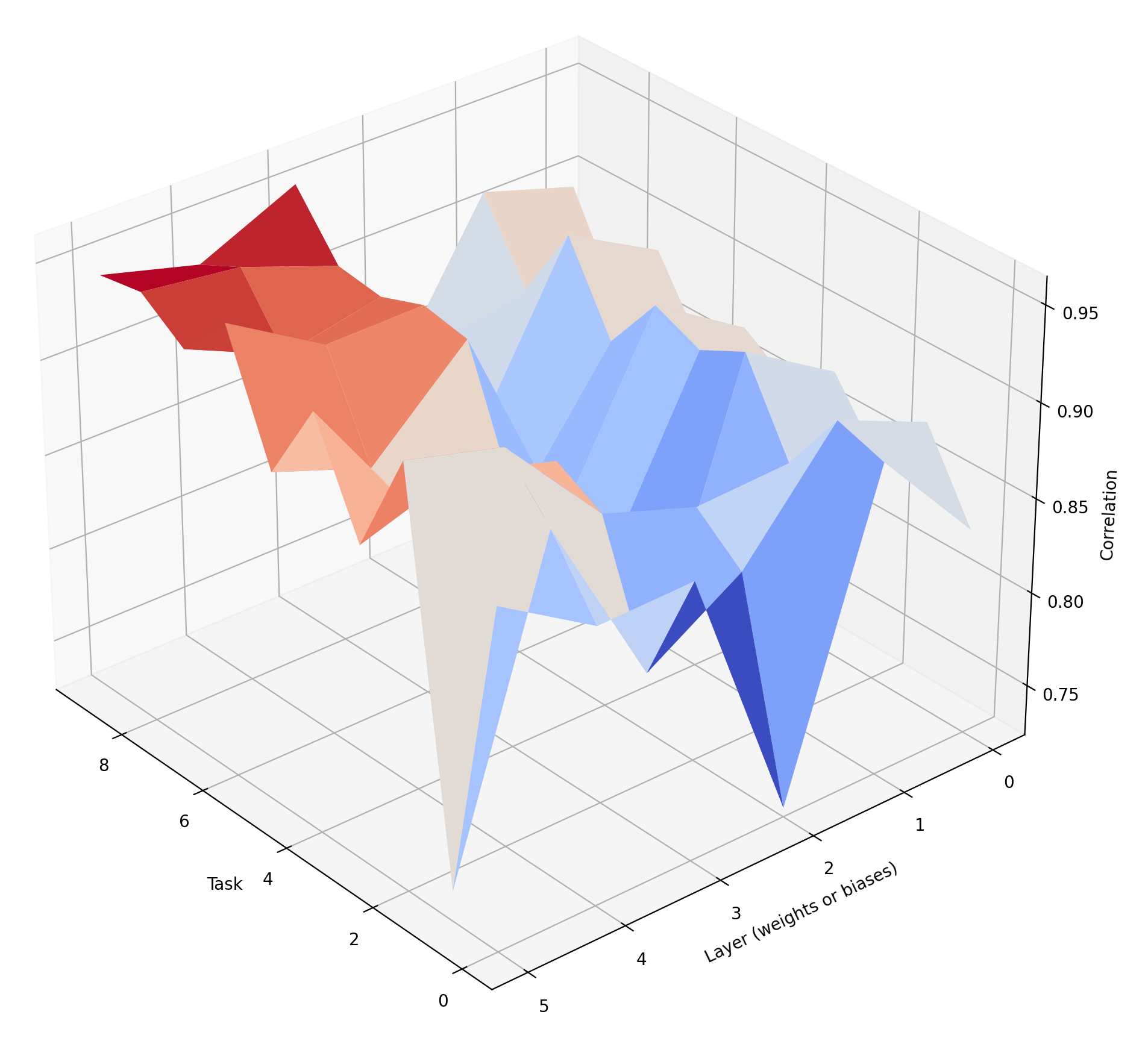}
\caption{\textit{Correlation between importances calculated by FIS and SI methods.}}
\label{figure:4}
\end{figure}

Consider Fig. \ref{figure:4} -- graph of correlations between importances calculated by FIS and SI methods.

Correlation between MAS and FIS importances is very high -- more than 0.75.

\subsection{Correlation between FIS and SIG importances}

\begin{figure}[h!]
\centering
\includegraphics[width=1\textwidth]{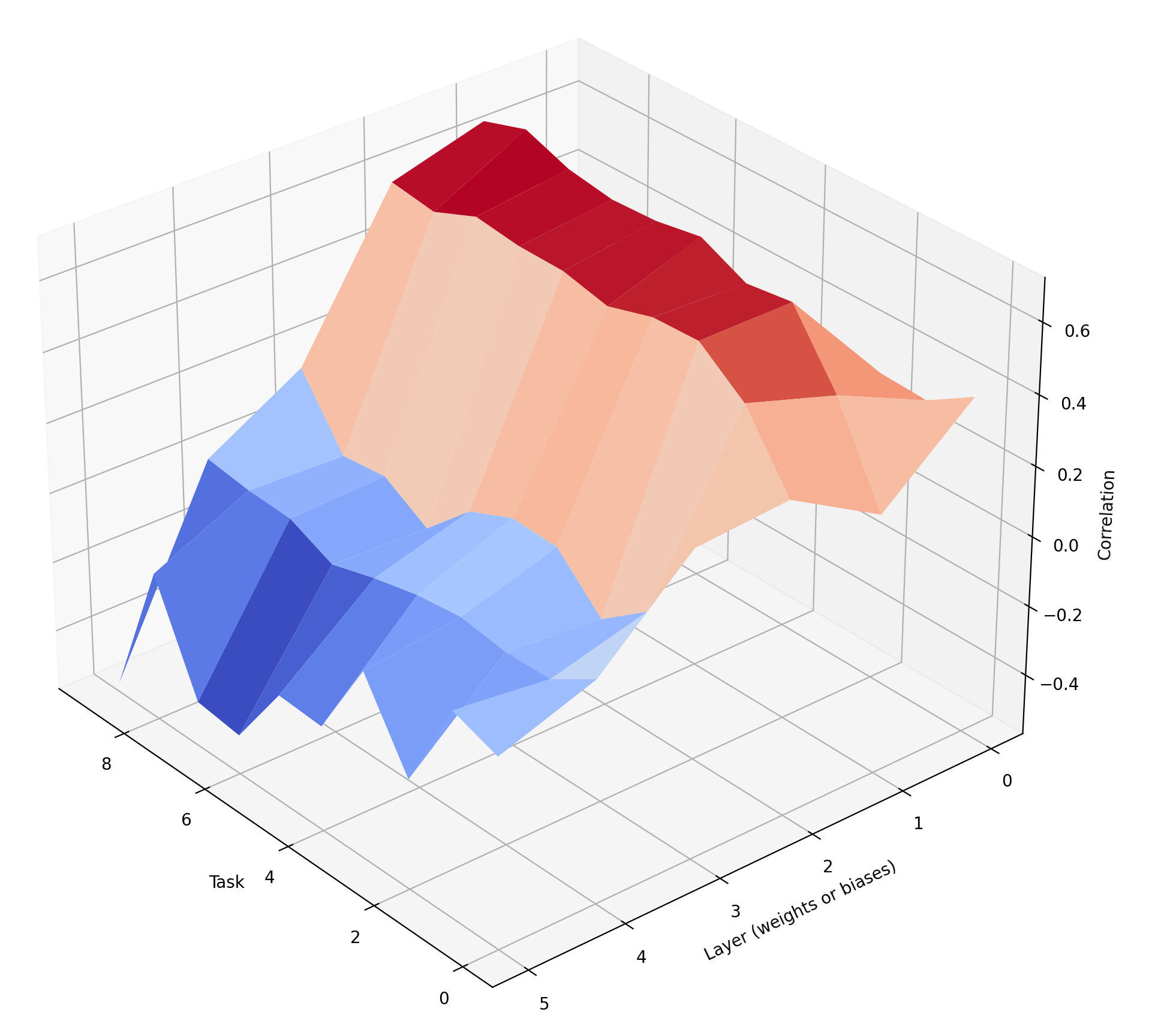}
\caption{\textit{Correlation between importances calculated by FIS and SIG methods.}}
\label{figure:5}
\end{figure}

Consider Fig. \ref{figure:5} -- graph of correlations between importances calculated by FIS and SIG methods.

Correlation between FIS and SIG importances is very different  -- from high (about 0.6) on input layers of neural network, to negative on output layer.

\subsection{Correlation between SI and SIG importances}

\begin{figure}[h!]
\centering
\includegraphics[width=1\textwidth]{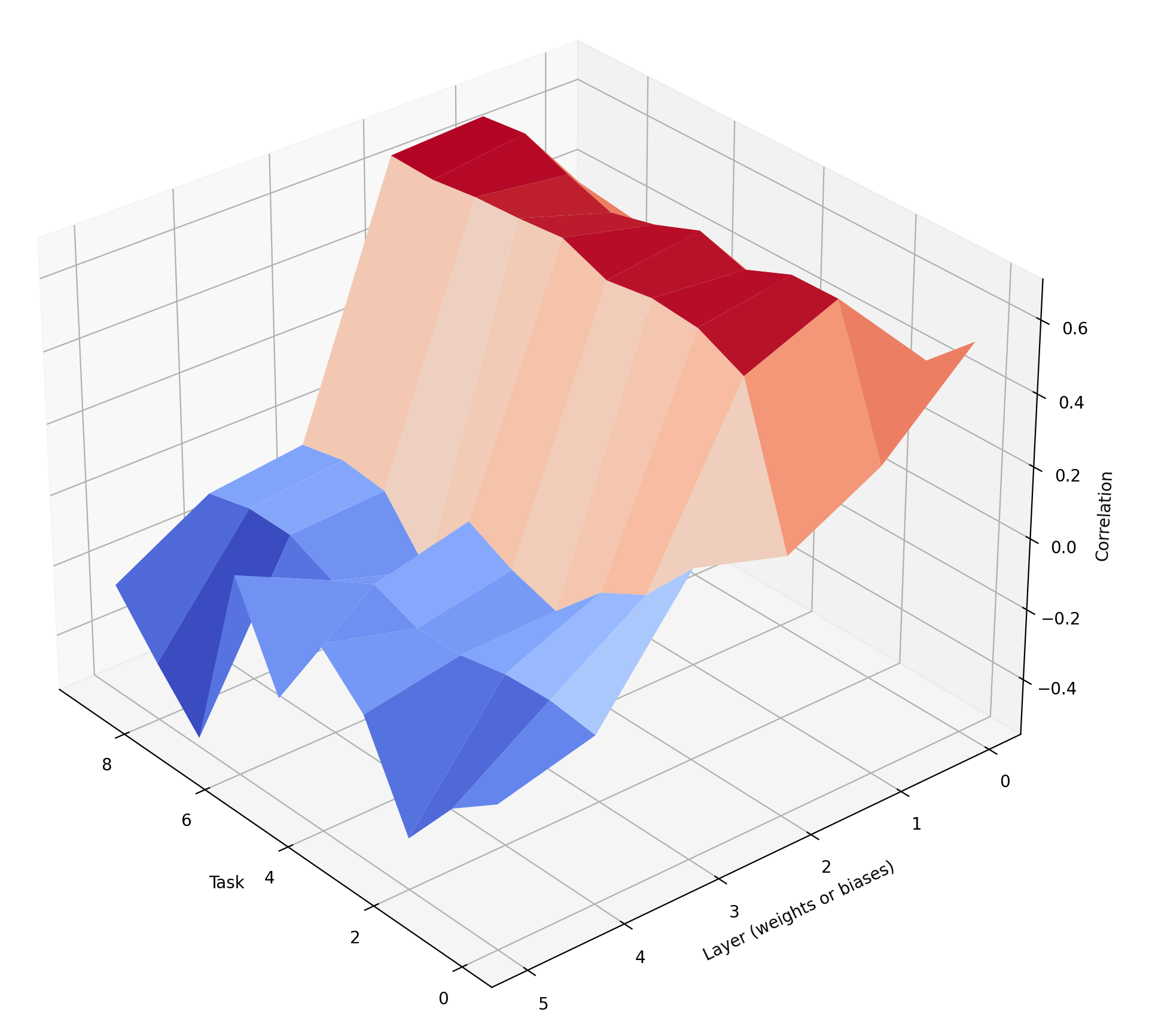}
\caption{\textit{Correlation between importances calculated by SI and SIG methods.}}
\label{figure:6}
\end{figure}

Consider Fig. \ref{figure:6} -- graph of correlations between importances calculated by SI and SIG methods.

Correlation between SI and SIG importances is very different  -- from high (about 0.6) on input layers of neural network, to negative on output layer.

\section{Discussion}

As you can see from the graphs, the methods for calculating the importances of MAS, FIS and SI produce approximately the same (highly correlated) importances of the neural network weights. And this explains their approximately the same efficiency in the EWC method. At the same time, the SIG method, which calculates the total passed signal, produces values that are very different (up to negative correlations) from the values using the MAS, FIS, and SI methods. As a result, a reasonable question arises: why, with such a strong difference from other types of importance, SIG importances allows us to get a comparable level of quality while overcoming the problem of catastrophic forgetting by EWC method?

If, as supposed in \cite{c10}, the long-term memory of real biological neurons operates similarly to SIG importances and the WVA method, i.e., neuronal plasticity decreases in proportion to the number of action potentials passed, then skill retention during sequential learning in such a mechanism will work better in shallow (1-2 layers) networks with fully connected layers, since they have the highest correlation with optimal importance (MAS) in terms of maintaining previously learned skills.


\end{document}